\documentclass[11pt]{article}
\usepackage[preprint]{acl}
\usepackage{times}
\usepackage{latexsym}
\usepackage[T1]{fontenc}
\usepackage[utf8]{inputenc}
\usepackage{microtype}
\usepackage{graphicx}
\usepackage{booktabs}
\usepackage{amsmath}
\usepackage{multirow}
\usepackage{xcolor}
\usepackage{float}
\usepackage{subcaption}
\usepackage{colortbl}
\usepackage{amsfonts}

\newcommand{\acc}[1]{%
  \ifdim #1 pt < 0.55pt
    \cellcolor[rgb]{0.957,0.71,0.706}#1%
  \else\ifdim #1 pt < 0.65pt
    \cellcolor[rgb]{0.976,0.859,0.718}#1%
  \else\ifdim #1 pt < 0.80pt
    \cellcolor[rgb]{1,1,0.718}#1%
  \else\ifdim #1 pt < 0.95pt
    \cellcolor[rgb]{0.776,0.918,0.718}#1%
  \else
    \cellcolor[rgb]{0.663,0.875,0.663}#1%
  \fi\fi\fi\fi
}

\graphicspath{{figures/}}

\title{No Memorization, No Detection: Output Distribution-Based Contamination Detection in Small Language Models}
\author{Omer Sela\\
Tel Aviv University}

\begin{document}
\maketitle

\begin{abstract}
CDD, or Contamination Detection via output Distribution, identifies data contamination by measuring the peakedness of a model's sampled outputs. We study the conditions under which this approach succeeds and fails on small language models ranging from 70M to 410M parameters. Using controlled contamination experiments on GSM8K, HumanEval, and MATH, we find that CDD's effectiveness depends critically on whether fine-tuning produces verbatim memorization. In the majority of conditions we test, CDD performs at chance level even when the data is verifiably contaminated and detectable by simpler methods. We show that probability-based methods, specifically perplexity and Min-k\% Prob, outperform CDD in all conditions where any method exceeds chance, suggesting that CDD's peakedness-based approach is insufficient for contamination detection in small language models. Our code is available at \url{https://github.com/Sela-Omer/Contamination-Detection-Small-LM}
\end{abstract}

\section{Introduction}

Data contamination, the presence of evaluation data in a model's training set, undermines the trustworthiness of language model benchmarks \citep{magar2022data, jacovi2023stop, sainz2023nlp}. As models are trained on increasingly large and opaque corpora, detecting contamination has become essential for reliable evaluation.

\citet{dong2024contamination} introduced CDD (Contamination Detection via output Distribution), which detects contamination by measuring the \emph{peakedness} of a model's output distribution. The intuition is that a model trained on specific data will produce abnormally similar outputs when sampled repeatedly on that data, because training collapses the output distribution toward the memorized answer. CDD requires only sampled text, making it applicable even to black-box models, and achieves 21--30\% relative improvement over baselines on 7B-parameter models.

A key question, however, is whether CDD's effectiveness depends on the fine-tuning regime. CDD detects peaked output distributions: cases where a model produces highly similar outputs regardless of sampling randomness. In practice, this occurs when training collapses the output distribution, typically through memorization. But contamination does not always produce such collapse. A model can be trained on leaked data and learn from it without converging to a single output, particularly when parameter-efficient methods limit the model's capacity.

We study this phenomenon systematically using Pythia models \citep{biderman2023pythia} (70M--410M parameters) fine-tuned on GSM8K \citep{cobbe2021training}, HumanEval \citep{chen2021evaluating}, and MATH \citep{hendrycks2021measuring} with controlled contamination levels. We vary three axes: model size, fine-tuning method (LoRA \citep{hu2022lora} with rank 8 and 256, and full fine-tuning), and training duration (3 and 20 epochs). Our main findings:

\begin{itemize}
    \item \textbf{CDD requires output distribution collapse to succeed.} With LoRA $r$=8 and 3 epochs, training loss decreases and all baseline methods confirm contamination, but CDD performs at chance because the model's output distribution has not collapsed.
    \item \textbf{A memorization threshold governs detectability.} CDD accuracy transitions sharply from chance to $>$90\% as fine-tuning capacity crosses a threshold. This threshold depends on the interaction of model size, adapter rank, and training duration.
    \item \textbf{Practical blind spot.} Parameter-efficient fine-tuning, increasingly the default for model adaptation, can produce contamination that CDD cannot detect. This represents a silent failure mode for peakedness-based detection.
\end{itemize}

\section{Related Work}

\paragraph{Data contamination detection.} The problem of benchmark data appearing in training corpora was first operationalized at scale by \citet{brown2020language}, who used 13-gram overlap to audit GPT-3's training set. Subsequent work developed detection methods that do not require training data access. \citet{li2023estimating} proposed perplexity as a proxy for memorization. \citet{shi2024detecting} introduced Min-k\% Prob, which flags text as seen during pretraining when its lowest-probability tokens are unusually high. \citet{golchin2024time} designed prompt-based probes using dataset metadata, and \citet{oren2024proving} proposed a statistical test with formal false-positive guarantees. \citet{deng2024investigating} combined retrieval-based auditing with slot-guessing probes.

CDD \citep{dong2024contamination} takes a distributional approach: it measures the peakedness of the edit distance distribution between a greedy reference output and temperature-sampled outputs. CDD was validated on 7B models (CodeLlama, Llama2, Bloom) using LoRA fine-tuning. Our work reveals that CDD's effectiveness is contingent on the fine-tuning regime producing sufficient memorization, a condition that does not always hold for smaller models or lower-rank adapters.

\paragraph{Memorization in language models.} \citet{carlini2023quantifying} showed that memorization increases with model scale and data duplication. \citet{tirumala2022memorization} found that larger models memorize faster and can memorize more data before overfitting signals emerge. \citet{biderman2023emergent} used the Pythia suite to study whether memorization can be predicted from early training checkpoints. These findings motivate our investigation: if memorization depends on model capacity, then detection methods that rely on memorization signals should also depend on capacity.

\paragraph{Parameter-efficient fine-tuning and memorization.} LoRA \citep{hu2022lora} adapts models by training low-rank matrices, updating only a small fraction of parameters. \citet{mireshghallah2022empirical} found that adapter-based fine-tuning can reduce susceptibility to extraction attacks compared to full fine-tuning. This is directly relevant to our central finding: LoRA's reduced parameter budget limits memorization, which in turn limits CDD's ability to detect contamination.

\section{Method}

\subsection{CDD: Contamination Detection via Output Distribution}

CDD \citep{dong2024contamination} rests on the observation that training alters a model's output distribution, making it more \emph{peaked} on data it has memorized. A model that has memorized a specific answer will tend to reproduce that answer even under stochastic sampling, whereas a model that has not memorized will produce diverse outputs. CDD operationalizes this intuition through four steps.

\paragraph{Sampling.} Given a prompt $x$ and a model $\mathcal{M}$, CDD generates one \emph{greedy} output $s_{t=0}$ using temperature $t$=0 (deterministic, always selecting the highest-probability token) and $n$ \emph{temperature samples} $S = \{s_1, \ldots, s_n\}$ at temperature $t$=0.8 (introducing controlled randomness). The greedy output serves as a reference: if the model has memorized the answer, it will be the memorized sequence. The temperature samples test whether the model can deviate from it. We use $n$=50, matching the original paper.

\paragraph{Edit distance computation.} CDD uses the token-level Levenshtein edit distance \citep{levenshtein1966binary} to measure similarity between generated sequences. Given two token sequences $a$ and $b$, the edit distance $\operatorname{ED}(a, b)$ is the minimum number of single-token insertions, deletions, and substitutions required to transform $a$ into $b$, computed via dynamic programming. CDD applies this in a star topology: the greedy reference $s_{t=0}$ is compared against each temperature sample $s_i$, rather than computing all pairwise distances. All sequences are tokenized using the model's BPE tokenizer and truncated to $l_{\max}$=100 tokens.

\paragraph{Peakedness.} The peakedness score measures what fraction of temperature samples are ``close'' to the greedy reference:
\begin{equation}
\text{Peak}(\mathcal{M}; x) = \frac{1}{n}\sum_{i=1}^{n} \mathbb{I}\big(\text{ED}(s_i, s_{t=0}) \leq \alpha \cdot l\big)
\end{equation}
where $l = \max\{|s| : s \in S \cup \{s_{t=0}\}\}$ is the maximum sequence length across all samples and $\alpha$=0.05 is a similarity threshold. With $l$=100 and $\alpha$=0.05, a sample counts as ``close'' if it differs from the greedy output by at most 5 token edits. High peakedness indicates that the model consistently reproduces similar outputs regardless of sampling randomness.

\paragraph{Classification.} A prompt is classified as contaminated if $\text{Peak}(\mathcal{M}; x) > \xi$, where $\xi$ is a decision threshold. The original paper uses a fixed $\xi$=0.01, calibrated on 7B models where contaminated examples produce high peakedness values. In our small-model setting, peakedness values are lower even when memorization occurs, so we select $\xi$ by maximizing the Youden index $J = \text{sensitivity} + \text{specificity} - 1$ \citep{youden1950index} over the evaluation set. This gives CDD every advantage and isolates the question of whether peakedness separates contaminated from clean examples at all.

\subsection{Baseline Detection Methods}

We compare CDD against three baselines that represent different detection paradigms:

\paragraph{N-gram overlap.} Following \citet{brown2020language}, we compute the 3-gram overlap between each test prompt and the training corpus. Since contaminated prompts are literally included in the training data, all of their n-grams appear in the corpus, yielding perfect separation from clean examples. This baseline requires access to the training corpus, which CDD does not, and serves primarily as ground-truth confirmation that the contamination injection succeeded rather than as a practical detection method.

\paragraph{Perplexity-based detection.} Following \citet{li2023estimating}, we compute the perplexity of each test prompt under the fine-tuned model. The intuition is that contaminated examples should have lower perplexity because the model has been trained on them. We classify examples with perplexity below an optimal threshold as contaminated. This baseline requires access to model output probabilities but not to the training corpus.

\paragraph{Min-k\% Prob.} Following \citet{shi2024detecting}, we compute the average log-probability of the $k$\% lowest-probability tokens in each test prompt under the fine-tuned model ($k$=20). The intuition is that contaminated text should have higher minimum token probabilities because even the least predictable tokens have been seen during training. We classify examples above an optimal threshold as contaminated. Like perplexity, this requires access to model output probabilities but not to the training corpus.


\subsection{Experimental Design}

\paragraph{Models.} We use three models from the Pythia suite \citep{biderman2023pythia}: Pythia-70M, Pythia-160M, and Pythia-410M. All are decoder-only transformers pre-trained on the Pile dataset under identical conditions, differing only in the number of parameters.

\paragraph{Datasets.} Our primary dataset is GSM8K \citep{cobbe2021training}, containing 7,473 grade-school math word problems with step-by-step solutions averaging 98 tokens. We randomly select 500 examples and split them into 300 for training, 100 for contamination injection, and 100 for evaluation. We also evaluate on two additional domains: HumanEval \citep{chen2021evaluating}, a code generation benchmark of 164 Python programming problems with canonical solutions averaging 69 tokens (split 100/32/32), and MATH \citep{hendrycks2021measuring}, a competition mathematics dataset with detailed solutions averaging 193 tokens (500 examples, split 300/100/100). For all datasets, the contamination set is repeated 0, 1, 5, or 10 times and concatenated with the training set. During fine-tuning, models see the full prompt-answer pair; during CDD detection, only the prompt is provided and the model generates continuations. Figures~\ref{fig:heatmap}--\ref{fig:peakedness} show GSM8K results; HumanEval and MATH results appear in Table~\ref{tab:baselines}.

\paragraph{Fine-tuning configurations.} We vary fine-tuning along two orthogonal dimensions to disentangle the effects of trainable capacity and training duration:
\begin{itemize}
    \item \textbf{Capacity}: LoRA \citep{hu2022lora} with rank $r$=8 (approximately 0.1--0.2\% of parameters trainable), LoRA with $r$=256 (approximately 4--6\%), and full fine-tuning (100\% of parameters).
    \item \textbf{Duration}: 3 training epochs and 20 training epochs.
\end{itemize}
Table~\ref{tab:params} shows the exact number of trainable parameters for each configuration. The range spans three orders of magnitude: from 98K (LoRA $r$=8 on 70M) to 405M (full fine-tuning on 410M). All configurations use learning rate $2 \times 10^{-4}$, batch size 8, gradient accumulation of 2 steps, and a warmup ratio of 0.1. LoRA adapters target the \texttt{query\_key\_value} projection in each attention layer. This yields 3 fine-tuning configurations $\times$ 3 model sizes $\times$ 4 contamination levels $\times$ (3 datasets on 3 epochs + 1 dataset on 20 epochs) = 144 experimental conditions, each trained and evaluated independently on 8 NVIDIA A100 GPUs. All code and configurations are available at \url{https://github.com/Sela-Omer/Contamination-Detection-Small-LM}.

\begin{table}[t]
\centering
\small
\begin{tabular}{lrrr}
\toprule
\textbf{Model} & \textbf{LoRA $r$=8} & \textbf{LoRA $r$=256} & \textbf{Full FT} \\
\midrule
70M   & 98K (0.14\%)  & 3.1M (4.3\%)  & 70.4M \\
160M  & 295K (0.18\%) & 9.4M (5.5\%)  & 162.3M \\
410M  & 786K (0.19\%) & 25.2M (5.9\%) & 405.3M \\
\bottomrule
\end{tabular}
\caption{Trainable parameters per configuration. The range spans three orders of magnitude, 98K -- 405M}
\label{tab:params}
\end{table}

\section{Results}

\subsection{The Core Finding: Contamination Without Memorization}

\begin{figure*}[t]
    \centering
    \includegraphics[width=\textwidth]{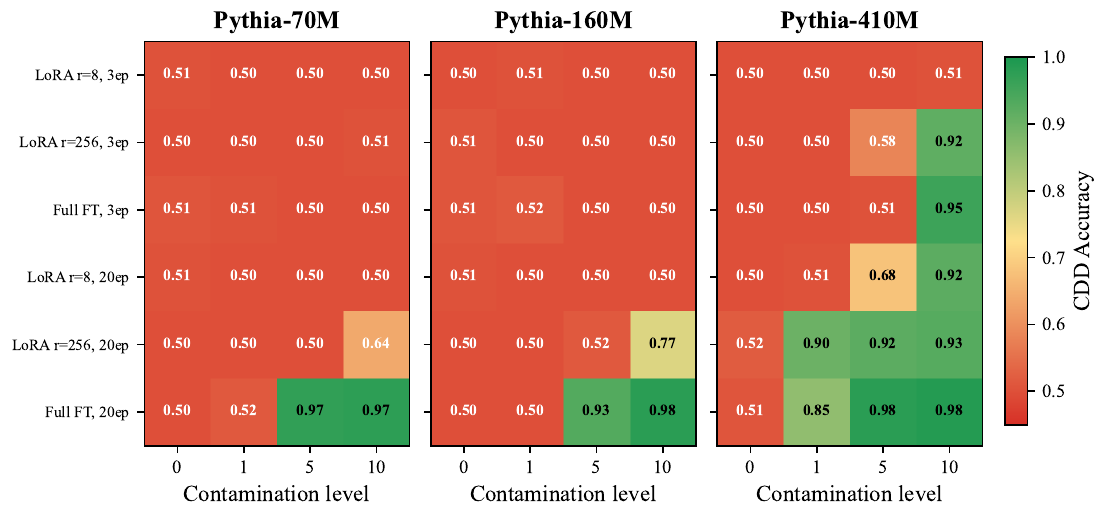}
    \caption{GSM8K: CDD detection accuracy across model sizes, fine-tuning methods, and contamination levels. Each cell shows accuracy (chance = 0.50). CDD fails entirely with low-capacity fine-tuning (top rows) but succeeds when fine-tuning produces output distribution collapse (bottom rows, larger models).}
    \label{fig:heatmap}
\end{figure*}

Our central result is shown in Figure~\ref{fig:heatmap}. On GSM8K, with LoRA $r$=8 and 3 training epochs, CDD achieves approximately 50\% accuracy (chance level) across \emph{all} model sizes and contamination levels, including the most extreme condition (410M, contamination level 10). Yet the training data is verifiably contaminated: all three baseline methods, including perplexity and Min-k\% Prob which require only model access, detect the contamination with accuracy well above chance (Table~\ref{tab:baselines}).

Despite being trained on the contaminated data, the model's output distribution does not collapse: temperature sampling produces diverse outputs, edit distances remain high, and peakedness stays at zero. We use \emph{output distribution collapse} to refer to the condition where a model produces highly similar outputs regardless of sampling randomness, operationally corresponding to high peakedness. In our qualitative analysis (Section~\ref{sec:qualitative}, Table~\ref{tab:qualitative}), we observe that this collapse coincides with the model reproducing the contaminated training data. The subsequent sections investigate the conditions under which collapse occurs.

\subsection{CDD vs. Baselines}

Table~\ref{tab:baselines} compares detection methods on Pythia-410M across contamination levels and all three datasets.

\paragraph{Probability-based methods outperform CDD.} The most striking pattern in Table~\ref{tab:baselines} is that whenever any detection method exceeds chance, perplexity and Min-k\% Prob outperform CDD. Defining above-chance as accuracy exceeding 0.55, CDD surpasses this threshold in only 7 of the 27 conditions shown, compared to 26 for perplexity and 25 for Min-k\%. The gap is largest precisely where it matters most: at low contamination levels and under parameter-efficient fine-tuning, where CDD is uniformly at chance but probability-based methods already show signal.

\paragraph{CDD's sensitivity to contamination level.} CDD requires heavy contamination to produce any signal. On GSM8K with LoRA $r$=256, CDD accuracy is 0.50 at $c$=1, 0.59 at $c$=5, and 0.92 at $c$=10. Perplexity already reaches 0.75 at $c$=1 and 1.00 at $c$=5. This means that at moderate contamination levels, CDD provides no useful signal while probability-based methods are already highly accurate.

\subsection{Detectability Threshold and Scale Effects}

CDD accuracy transitions sharply from chance to strong detection as fine-tuning capacity increases. On GSM8K, for Pythia-410M at contamination level 10, the jump from LoRA $r$=8 to $r$=256 (both at 3 epochs) takes accuracy from chance to 0.915, demonstrating that the number of trainable parameters is the primary driver (Table~\ref{tab:baselines}). Extended training can partially compensate for low rank: LoRA $r$=8 at 20 epochs reaches 0.920, comparable to LoRA $r$=256 at 3 epochs. Full fine-tuning achieves 0.955 with only 3 epochs, and 0.985 with 20 epochs. The full results across all conditions are shown in Figure~\ref{fig:heatmap}.

Model size amplifies this effect, but only above the detectability threshold. On GSM8K with LoRA $r$=256 at 20 epochs and contamination level 10: 70M achieves 0.640, 160M achieves 0.765, and 410M achieves 0.925. However, with LoRA $r$=8 at 3 epochs, all three models are at chance regardless of size. Scale amplifies output distribution collapse but cannot create it when the fine-tuning capacity is insufficient.

\begin{table*}[t]
\centering
\small
\begin{tabular}{ll cccc cccc cccc}
\toprule
& & \multicolumn{4}{c}{\textbf{GSM8K}} & \multicolumn{4}{c}{\textbf{HumanEval}} & \multicolumn{4}{c}{\textbf{MATH}} \\
\cmidrule(lr){3-6} \cmidrule(lr){7-10} \cmidrule(lr){11-14}
\textbf{FT} & \textbf{$c$} & \textbf{CDD} & \textbf{PPL} & \textbf{Mk\%} & \textbf{Ng} & \textbf{CDD} & \textbf{PPL} & \textbf{Mk\%} & \textbf{Ng} & \textbf{CDD} & \textbf{PPL} & \textbf{Mk\%} & \textbf{Ng} \\
\midrule
\multirow{3}{*}{\rotatebox{90}{\scriptsize LoRA 8}}
 & 1  & \acc{.50} & \acc{.58} & \acc{.60} & \acc{1.0} & \acc{.53} & \acc{.53} & \acc{.53} & \acc{1.0} & \acc{.50} & \acc{.58} & \acc{.60} & \acc{1.0} \\
 & 5  & \acc{.50} & \acc{.63} & \acc{.63} & \acc{1.0} & \acc{.50} & \acc{.56} & \acc{.55} & \acc{1.0} & \acc{.51} & \acc{.64} & \acc{.66} & \acc{1.0} \\
 & 10 & \acc{.51} & \acc{.78} & \acc{.77} & \acc{1.0} & \acc{.50} & \acc{.67} & \acc{.70} & \acc{1.0} & \acc{.50} & \acc{.73} & \acc{.77} & \acc{1.0} \\
\midrule
\multirow{3}{*}{\rotatebox{90}{\scriptsize LoRA 256}}
 & 1  & \acc{.50} & \acc{.75} & \acc{.74} & \acc{1.0} & \acc{.50} & \acc{.59} & \acc{.64} & \acc{1.0} & \acc{.51} & \acc{.64} & \acc{.67} & \acc{1.0} \\
 & 5  & \acc{.59} & \acc{1.0} & \acc{1.0} & \acc{1.0} & \acc{.53} & \acc{.97} & \acc{1.0} & \acc{1.0} & \acc{.53} & \acc{.95} & \acc{.96} & \acc{1.0} \\
 & 10 & \acc{.92} & \acc{1.0} & \acc{1.0} & \acc{1.0} & \acc{.84} & \acc{1.0} & \acc{1.0} & \acc{1.0} & \acc{.83} & \acc{1.0} & \acc{1.0} & \acc{1.0} \\
\midrule
\multirow{3}{*}{\rotatebox{90}{\scriptsize Full FT}}
 & 1  & \acc{.50} & \acc{1.0} & \acc{1.0} & \acc{1.0} & \acc{.52} & \acc{.64} & \acc{.73} & \acc{1.0} & \acc{.50} & \acc{.77} & \acc{.81} & \acc{1.0} \\
 & 5  & \acc{.51} & \acc{1.0} & \acc{1.0} & \acc{1.0} & \acc{.58} & \acc{1.0} & \acc{1.0} & \acc{1.0} & \acc{.51} & \acc{.98} & \acc{.97} & \acc{1.0} \\
 & 10 & \acc{.96} & \acc{1.0} & \acc{1.0} & \acc{1.0} & \acc{.72} & \acc{1.0} & \acc{1.0} & \acc{1.0} & \acc{.51} & \acc{.98} & \acc{.99} & \acc{1.0} \\
\bottomrule
\end{tabular}
\caption{Detection accuracy on Pythia-410M across contamination levels $c \in \{1, 5, 10\}$ and three datasets (3 training epochs). CDD requires only sampled text; PPL and Min-k\% Prob (Mk\%) \citep{shi2024detecting} require output probabilities; N-gram (Ng) requires training corpus access. Probability-based methods detect contamination at lower levels and in conditions where CDD fails, across all three domains.}
\label{tab:baselines}
\end{table*}

\subsection{Training Loss Does Not Predict Detectability}

\begin{figure}[t]
    \centering
    \includegraphics[width=\columnwidth]{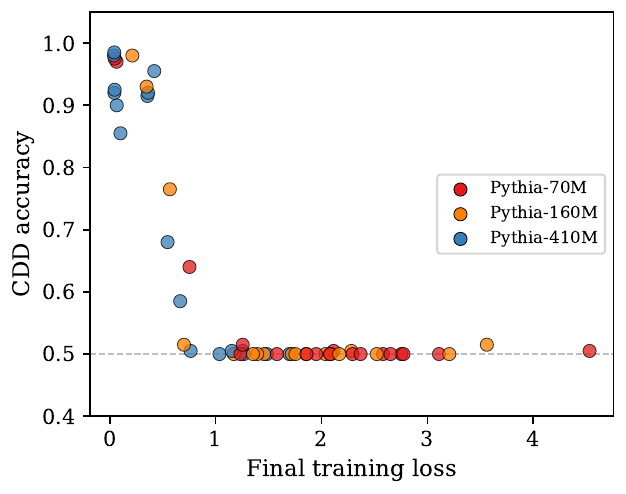}
    \caption{GSM8K: Final training loss vs. CDD accuracy across all contaminated conditions. Each point is one model/ft-method/contamination-level combination. Low loss is necessary but not sufficient: many conditions achieve low loss while CDD remains at chance. CDD accuracy rises only when loss approaches zero, indicating output distribution collapse.}
    \label{fig:loss}
\end{figure}

Figure~\ref{fig:loss} shows the relationship between final training loss and CDD accuracy across all contaminated GSM8K conditions. Two distinct regimes emerge. In the ``learning without memorization'' regime (loss between 1.0 and 3.0), models have learned from the contaminated data, as evidenced by decreasing loss, but CDD accuracy remains at chance. In the ``memorization'' regime (loss below approximately 0.5), models have memorized specific training examples and CDD accuracy rises sharply. The transition between these regimes is abrupt, not gradual: there is no intermediate zone where CDD partially works. This confirms that CDD responds to output distribution collapse, not to learning per se.

\subsection{Peakedness Distributions}

\begin{figure}[t]
    \centering
    \includegraphics[width=\columnwidth]{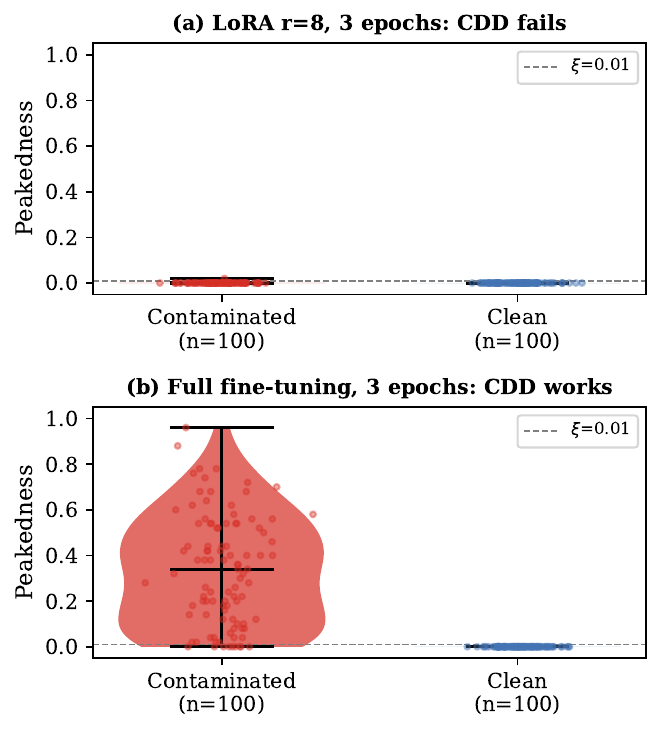}
    \caption{GSM8K: Peakedness distributions for Pythia-410M at contamination level 10. (a) With LoRA $r$=8, both contaminated and clean examples cluster at zero. (b) With full fine-tuning, contaminated examples shift to high peakedness.}
    \label{fig:peakedness}
\end{figure}

Figure~\ref{fig:peakedness} contrasts the peakedness distributions under two fine-tuning regimes. When CDD fails (LoRA $r$=8, 3 epochs), both contaminated and clean examples have peakedness concentrated at zero, making them indistinguishable. When CDD succeeds (full fine-tuning, 3 epochs), contaminated examples shift to high peakedness ($\mu$=0.34) while clean examples remain at zero, providing clear separation.

\subsection{Qualitative Analysis}
\label{sec:qualitative}

Table~\ref{tab:qualitative} illustrates the difference between learning and output distribution collapse on a contaminated GSM8K example. With LoRA $r$=8 (3 epochs), the model has learned the GSM8K format (step-by-step arithmetic with \texttt{<<>>} annotations) but produces \emph{different} outputs each time it is sampled. The greedy output and the temperature sample share the general structure but diverge in content, resulting in high edit distance and zero peakedness. With full fine-tuning (3 epochs), the model's output distribution has collapsed: both the greedy output and the temperature sample produce the same sequence. The edit distance is near zero and peakedness is high, allowing CDD to detect the contamination.

This example captures the core distinction: the LoRA model is contaminated (it was trained on this example 10 times) but has not memorized it. The fully fine-tuned model is contaminated \emph{and} has memorized it. CDD can only detect the latter.

\begin{table*}[t]
\centering
\small
\begin{tabular}{p{1.8cm}p{13.5cm}}
\toprule
\textbf{Prompt} & \textit{Question: Martha has 18 crayons. She lost half of them, so she bought a new set of 20 crayons. How many crayons in total does Martha have after the purchase? Answer:} \\
\textbf{Ground truth} & In the beginning, Martha lost 18 / 2 = \texttt{<<18/2=9>>}9 crayons. So after she bought a set of new crayons, she has in total 9 + 20 = \texttt{<<9+20=29>>}29 crayons. \#\#\#\# 29 \\
\midrule
\multicolumn{2}{l}{\textbf{LoRA $r$=8, 3 epochs} (CDD: 0.505, PPL: 0.78, Min-k\%: 0.77)} \\
\textbf{Greedy} & Martha lost half of her set of 18 crayons because she lost half of her set of 18 crayons. She bought a new set of 20 crayons because she bought 20 crayons. So, Martha has 18 + 20 = \texttt{<<18+20=42>>}42 crayons. \#\#\#\# \textbf{42} \\
\textbf{Sample} & A pair of crayons has 18 / 2 = \texttt{<<18/2=9>>}9 crayons. Each new set has 20 + 9 = \texttt{<<20+9=38>>}38 crayons. \#\#\#\# \textbf{38} \\
\midrule
\multicolumn{2}{l}{\textbf{Full fine-tuning, 3 epochs} (CDD: 0.955, PPL: 1.00, Min-k\%: 1.00)} \\
\textbf{Greedy} & In the beginning, Martha lost 18 / 2 = \texttt{<<18/2=9>>}9 crayons. So after she bought a set of new crayons, she has in total 9 + 20 = \texttt{<<9+20=29>>}29 crayons. \#\#\#\# \textbf{29} \\
\textbf{Sample} & In the beginning, Martha lost 18 / 2 = \texttt{<<18/2=9>>}9 crayons. So after she bought a set of new crayons, she has in total 9 + 20 = \texttt{<<9+20=29>>}29 crayons. \#\#\#\# \textbf{29} \\
\bottomrule
\end{tabular}
\caption{Model outputs on a contaminated GSM8K example (Pythia-410M, contamination level 10). With LoRA $r$=8, the model produces different wrong answers each time (high edit distance, zero peakedness). With full fine-tuning, it reproduces the ground truth verbatim (near-zero edit distance, high peakedness). Both models were trained on this example 10 times. CDD accuracy shown is for the full condition, not this single example.}
\label{tab:qualitative}
\end{table*}

\paragraph{The memorization threshold across domains.} The LoRA $r$=256 results are particularly informative because this configuration sits near the memorization threshold across all three datasets. At $c$=10 on 410M, LoRA $r$=256 achieves CDD accuracy of 0.92 (GSM8K), 0.84 (HumanEval), and 0.83 (MATH). These are the conditions where CDD works best with parameter-efficient fine-tuning. Yet at $c$=1, the same configuration drops to chance on all three datasets (0.50, 0.50, 0.51), while perplexity already achieves 0.75, 0.59, and 0.64 respectively.

\paragraph{The low-contamination regime.} The $c$=1 rows of Table~\ref{tab:baselines} deserve particular attention because single-repetition contamination is the most realistic scenario. In all 9 conditions at $c$=1 (3 FT methods $\times$ 3 datasets), CDD is at or near chance (0.50--0.53). Perplexity and Min-k\% show meaningful signal in 6 of 9 conditions, with accuracy reaching 1.00 for full fine-tuning on GSM8K. This means that in the most practically relevant contamination scenario, CDD provides zero information while probability-based methods are already useful.

\subsection{CDD Hyperparameter Sensitivity}

To verify that CDD's failure is not an artifact of the default hyperparameter settings, we ablate the peakedness threshold $\alpha$, sampling temperature $t$, and number of samples $n$ on GSM8K at $c$=10 across all model sizes and fine-tuning methods (Table~\ref{tab:ablation}). No hyperparameter setting rescues CDD when it fails: for 70M and 160M, CDD remains at chance regardless of settings, for all fine-tuning methods. For 410M with LoRA $r$=8, CDD stays at chance across all 14 hyperparameter values tested. Tuning helps only where CDD already works (410M with lora256/full): lower temperature and higher $\alpha$ improve accuracy modestly, but these gains are small compared to the gap between CDD and probability-based methods.

\begin{table*}[t]
\centering
\small
\begin{tabular}{lll ccccc ccccc cccc}
\toprule
& & & \multicolumn{5}{c}{\textbf{Peakedness $\alpha$}} & \multicolumn{5}{c}{\textbf{Temperature $t$}} & \multicolumn{4}{c}{\textbf{Samples $n$}} \\
\cmidrule(lr){4-8} \cmidrule(lr){9-13} \cmidrule(lr){14-17}
\textbf{FT} & \textbf{Model} & & .01 & .02 & \underline{.05} & .10 & .20 & .4 & .6 & \underline{.8} & 1.0 & 1.2 & 10 & 25 & \underline{50} & 100 \\
\midrule
\multirow{3}{*}{\rotatebox{90}{\scriptsize LoRA 8}}
 & 70M  && \acc{.50} & \acc{.50} & \acc{.50} & \acc{.51} & \acc{.50} & \acc{.54} & \acc{.50} & \acc{.50} & \acc{.50} & \acc{.50} & \acc{.50} & \acc{.50} & \acc{.50} & \acc{.50} \\
 & 160M && \acc{.50} & \acc{.50} & \acc{.50} & \acc{.50} & \acc{.50} & \acc{.52} & \acc{.51} & \acc{.50} & \acc{.50} & \acc{.50} & \acc{.50} & \acc{.50} & \acc{.50} & \acc{.50} \\
 & 410M && \acc{.50} & \acc{.50} & \acc{.51} & \acc{.51} & \acc{.52} & \acc{.53} & \acc{.50} & \acc{.51} & \acc{.50} & \acc{.50} & \acc{.50} & \acc{.50} & \acc{.51} & \acc{.51} \\
\midrule
\multirow{3}{*}{\rotatebox{90}{\scriptsize LoRA 256}}
 & 70M  && \acc{.50} & \acc{.50} & \acc{.51} & \acc{.50} & \acc{.50} & \acc{.51} & \acc{.51} & \acc{.51} & \acc{.50} & \acc{.50} & \acc{.51} & \acc{.51} & \acc{.51} & \acc{.51} \\
 & 160M && \acc{.50} & \acc{.50} & \acc{.50} & \acc{.50} & \acc{.50} & \acc{.50} & \acc{.50} & \acc{.50} & \acc{.50} & \acc{.50} & \acc{.50} & \acc{.50} & \acc{.50} & \acc{.50} \\
 & 410M && \acc{.89} & \acc{.90} & \acc{.92} & \acc{.95} & \acc{.95} & \acc{.95} & \acc{.93} & \acc{.92} & \acc{.89} & \acc{.80} & \acc{.88} & \acc{.90} & \acc{.92} & \acc{.93} \\
\midrule
\multirow{3}{*}{\rotatebox{90}{\scriptsize Full FT}}
 & 70M  && \acc{.50} & \acc{.50} & \acc{.50} & \acc{.52} & \acc{.55} & \acc{.59} & \acc{.54} & \acc{.50} & \acc{.51} & \acc{.50} & \acc{.50} & \acc{.50} & \acc{.50} & \acc{.52} \\
 & 160M && \acc{.50} & \acc{.50} & \acc{.50} & \acc{.50} & \acc{.50} & \acc{.52} & \acc{.51} & \acc{.50} & \acc{.50} & \acc{.50} & \acc{.50} & \acc{.50} & \acc{.50} & \acc{.50} \\
 & 410M && \acc{.91} & \acc{.94} & \acc{.96} & \acc{.96} & \acc{.98} & \acc{.97} & \acc{.96} & \acc{.96} & \acc{.91} & \acc{.72} & \acc{.91} & \acc{.94} & \acc{.96} & \acc{.97} \\
\bottomrule
\end{tabular}
\caption{CDD hyperparameter sensitivity (GSM8K, $c$=10, 3 epochs). Default settings used throughout the paper are \underline{underlined}: $\alpha$=0.05, $t$=0.8, $n$=50. Only 410M with LoRA $r$=256 or full FT shows any sensitivity to hyperparameters. All other conditions remain at chance regardless of settings.}
\label{tab:ablation}
\end{table*}

\subsection{Stability Across Data Splits}

To verify that our results are not artifacts of a particular data split, we repeat the experiment with 8 additional random 300/100/100 splits of GSM8K (Pythia-410M, $c$=10, 3 epochs), for a total of 9 independent runs including the original. Table~\ref{tab:variance} reports mean $\pm$ standard deviation across runs. CDD's failure under LoRA $r$=8 is highly stable (0.502 $\pm$ 0.003, never exceeding 0.51), confirming it is at chance regardless of which examples are selected. CDD's success under LoRA $r$=256 is also consistent (0.924 $\pm$ 0.017). Probability-based methods show low variance: perplexity achieves 0.757 $\pm$ 0.034 under LoRA $r$=8 and perfect accuracy under LoRA $r$=256 across all 9 runs.

\begin{table}[t]
\centering
\small
\begin{tabular}{l cc}
\toprule
& \textbf{LoRA $r$=8} & \textbf{LoRA $r$=256} \\
\midrule
CDD     & .50 $\pm$ .003 & .92 $\pm$ .017 \\
PPL     & .76 $\pm$ .034 & 1.0 $\pm$ .000 \\
Min-k\% & .74 $\pm$ .027 & 1.0 $\pm$ .000 \\
N-gram  & 1.0 $\pm$ .000 & 1.0 $\pm$ .000 \\
\bottomrule
\end{tabular}
\caption{Detection accuracy (mean $\pm$ std) across 9 independent data splits (GSM8K, 410M, $c$=10, 3 epochs).}
\label{tab:variance}
\end{table}

\section{Discussion}

\paragraph{Reconciling with the original CDD results.} The original CDD paper also uses LoRA but on 7B-parameter models, where even LoRA provides millions of trainable parameters. LoRA $r$=8 on a 7B model yields roughly 4M trainable parameters; the same rank on our 70M model yields only 98K. Our LoRA $r$=256, which provides 3--25M trainable parameters, is closer in absolute capacity to what low-rank LoRA provides on 7B models, and this is where CDD begins to work in our experiments. The relevant factor is not the LoRA rank itself but the absolute number of trainable parameters.

\paragraph{The relationship between output collapse and detectability.} CDD measures peakedness: the fraction of temperature samples that are close to the greedy output. By definition, high peakedness means the model produces consistent outputs. The key empirical finding is that in our setting, this consistency only arises when the model has memorized the contaminated training data and reproduces it. In the qualitative examples (Table~\ref{tab:qualitative}), the collapsed outputs match the ground-truth training sequences. When models learn from contaminated data without memorizing it, their outputs remain diverse and CDD cannot distinguish them from clean examples, even though probability-based methods detect the contamination.

\paragraph{Variation across domains.} CDD's accuracy varies across datasets at the same fine-tuning configuration: full fine-tuning on 410M at $c$=10 achieves 0.96 on GSM8K but 0.72 on HumanEval and 0.51 on MATH. The reasons for this variation are not fully clear and may involve differences in answer length, structural regularity, or vocabulary diversity. What is clear is that this variation does not affect the probability-based baselines, which achieve near-perfect accuracy across all three domains. This reinforces that CDD's sensitivity to domain-specific factors is an additional practical limitation beyond the memorization threshold.

\paragraph{Why probability-based methods succeed where CDD fails.} The consistent superiority of perplexity and Min-k\% Prob across all conditions deserves explanation. These methods evaluate the model's \emph{internal} probability distribution over the prompt tokens, which shifts even when the model has not memorized specific outputs. Fine-tuning on contaminated data reduces the model's surprise at seeing those prompts, lowering perplexity and raising minimum token probabilities, even with LoRA $r$=8. CDD, by contrast, requires the model's \emph{external} behavior (generated text) to change in a specific way: the output distribution must collapse toward a single point. This is a much stronger condition that requires the model to converge to consistent outputs, not just to become familiar with the data.

\paragraph{The role of training loss.} Training loss decreases in all conditions, including those where CDD fails. For 410M with LoRA $r$=8 at 3 epochs, loss drops from 2.35 to 1.26, yet CDD accuracy is 0.505. Loss reduction reflects learning, but not necessarily memorization. The gap between ``the model learned something from this data'' and ``the model memorized this data'' is precisely where CDD's blind spot lies, but it is not a blind spot for all detection methods.

\paragraph{Implications for practitioners.} Our results suggest that CDD should not be used as the sole contamination detection method for small language models, particularly when parameter-efficient fine-tuning is involved. In the increasingly common scenario where models are adapted with LoRA, CDD may provide false assurance that no contamination exists. Perplexity and Min-k\% Prob, which require access to output probabilities rather than just sampled text, outperform CDD in all conditions where any method exceeds chance.

\paragraph{Scope of our claims about sample-based detection.} CDD represents one specific sample-based strategy: it reduces the output distribution to a single scalar (peakedness) via token-level edit distance to a greedy reference, so it can only detect contamination that manifests as near-verbatim reproduction. Alternative two-sample tests operate differently: the MMD \citep{gretton2012kernel} compares distributions by measuring the largest difference over functions in a reproducing kernel Hilbert space, while classifier two-sample tests \citep{lopez2017revisiting} train a binary classifier to distinguish two sets of samples and flag a difference whenever accuracy exceeds chance. Both are sensitive to distributional shifts beyond single-mode collapse. However, applying these tests to variable-length text generations requires nontrivial design choices, including a representation space (e.g., sentence embeddings) and a reference distribution to compare against, making them not a drop-in replacement for CDD. Whether such approaches can detect contamination in the ``learning without memorization'' regime we identify remains open.

\section{Conclusion}

We present evidence that CDD is unreliable for contamination detection on small language models, succeeding only when training produces memorization strong enough to collapse the output distribution. This finding holds across reasoning (GSM8K), code generation (HumanEval), and competition mathematics (MATH). When fine-tuning does not produce memorization, CDD fails silently, performing at chance level even on heavily contaminated data. Detection accuracy depends on the interaction of model size, fine-tuning capacity, and training duration, with a sharp memorization threshold governing the transition from undetectable to detectable contamination. Probability-based methods, specifically perplexity and Min-k\% Prob, outperform CDD in all conditions where any method exceeds chance, including those where CDD fails entirely. Our results suggest that CDD's peakedness-based approach is insufficient for contamination detection in small language models, and that the community should consider probability-based alternatives when auditing models at this scale.

\section{Limitations} 
Our study is limited to the Pythia family at 70M--410M parameters. The original CDD paper demonstrated strong results on 7B models, where even LoRA provides millions of trainable parameters; our findings are specific to the small-model regime and should not be extrapolated to larger scales without further investigation.

Our contamination setup injects repeated examples into fine-tuning data; pre-training contamination, where benchmark data appears in the original training corpus without explicit repetition, may produce different dynamics.

We use relatively small datasets (500 GSM8K, 164 HumanEval, 500 MATH), and the memorization threshold we observe may shift with larger datasets where each example receives less training signal.

Our conclusions about sample-based detection are specific to CDD's peakedness statistic; as discussed in Section~5, alternative sample-based tests that are sensitive to broader distributional properties may behave differently.

\bibliography{paper}

\onecolumn
\appendix

\section{Threshold-Free Metrics}
\label{app:auroc}

Table~\ref{tab:auroc} reports AUROC and AUPRC for all detection methods across model sizes, fine-tuning methods, and datasets at contamination level 10. These threshold-free metrics evaluate the discriminative quality of each method's raw scores without committing to any classification threshold. The results confirm the accuracy-based findings from the main paper: probability-based methods consistently outperform CDD across all conditions.

\begin{table*}[htbp]
\centering
\small
\begin{tabular}{lll cccc cccc}
\toprule
& & & \multicolumn{4}{c}{\textbf{AUROC}} & \multicolumn{4}{c}{\textbf{AUPRC}} \\
\cmidrule(lr){4-7} \cmidrule(lr){8-11}
\textbf{Dataset} & \textbf{FT} & \textbf{Model} & \textbf{CDD} & \textbf{PPL} & \textbf{Mk\%} & \textbf{Ng} & \textbf{CDD} & \textbf{PPL} & \textbf{Mk\%} & \textbf{Ng} \\
\midrule
\multirow{9}{*}{GSM8K}
 & \multirow{3}{*}{r=8}   & 70M  & \acc{.50} & \acc{.58} & \acc{.57} & \acc{1.0} & \acc{.50} & \acc{.63} & \acc{.62} & \acc{1.0} \\
 &                         & 160M & \acc{.50} & \acc{.67} & \acc{.65} & \acc{1.0} & \acc{.50} & \acc{.70} & \acc{.67} & \acc{1.0} \\
 &                         & 410M & \acc{.50} & \acc{.86} & \acc{.83} & \acc{1.0} & \acc{.50} & \acc{.86} & \acc{.84} & \acc{1.0} \\
\cmidrule(lr){2-11}
 & \multirow{3}{*}{r=256}  & 70M  & \acc{.50} & \acc{.87} & \acc{.83} & \acc{1.0} & \acc{.50} & \acc{.88} & \acc{.85} & \acc{1.0} \\
 &                         & 160M & \acc{.50} & \acc{.98} & \acc{.96} & \acc{1.0} & \acc{.50} & \acc{.98} & \acc{.96} & \acc{1.0} \\
 &                         & 410M & \acc{.93} & \acc{1.0} & \acc{1.0} & \acc{1.0} & \acc{.94} & \acc{1.0} & \acc{1.0} & \acc{1.0} \\
\cmidrule(lr){2-11}
 & \multirow{3}{*}{Full}   & 70M  & \acc{.50} & \acc{1.0} & \acc{1.0} & \acc{1.0} & \acc{.50} & \acc{1.0} & \acc{1.0} & \acc{1.0} \\
 &                         & 160M & \acc{.50} & \acc{1.0} & \acc{1.0} & \acc{1.0} & \acc{.50} & \acc{1.0} & \acc{1.0} & \acc{1.0} \\
 &                         & 410M & \acc{.96} & \acc{1.0} & \acc{1.0} & \acc{1.0} & \acc{.96} & \acc{1.0} & \acc{1.0} & \acc{1.0} \\
\midrule
\multirow{3}{*}{HumanEval}
 & r=8   & 410M & \acc{.50} & \acc{.66} & \acc{.71} & \acc{1.0} & \acc{.50} & \acc{.64} & \acc{.67} & \acc{1.0} \\
 & r=256 & 410M & \acc{.84} & \acc{1.0} & \acc{1.0} & \acc{1.0} & \acc{.84} & \acc{1.0} & \acc{1.0} & \acc{1.0} \\
 & Full  & 410M & \acc{.72} & \acc{1.0} & \acc{1.0} & \acc{1.0} & \acc{.72} & \acc{1.0} & \acc{1.0} & \acc{1.0} \\
\midrule
\multirow{3}{*}{MATH}
 & r=8   & 410M & \acc{.50} & \acc{.78} & \acc{.81} & \acc{1.0} & \acc{.50} & \acc{.74} & \acc{.78} & \acc{1.0} \\
 & r=256 & 410M & \acc{.85} & \acc{1.0} & \acc{1.0} & \acc{1.0} & \acc{.86} & \acc{1.0} & \acc{1.0} & \acc{1.0} \\
 & Full  & 410M & \acc{.50} & \acc{1.0} & \acc{1.0} & \acc{1.0} & \acc{.50} & \acc{1.0} & \acc{1.0} & \acc{1.0} \\
\bottomrule
\end{tabular}
\caption{AUROC and AUPRC at $c$=10 (3 epochs). Threshold-free metrics confirm the accuracy-based findings.}
\label{tab:auroc}
\end{table*}

\section{Per-Dataset Detailed Results}
\label{app:datasets}

This section presents the full CDD analysis for HumanEval and MATH, mirroring the GSM8K figures in the main paper. For each dataset, we show: (1) a heatmap of CDD accuracy across all conditions; (2) the relationship between training loss and CDD accuracy; and (3) peakedness distributions contrasting failing and succeeding conditions.

\begin{figure*}[htbp]
    \centering
    \includegraphics[width=\textwidth]{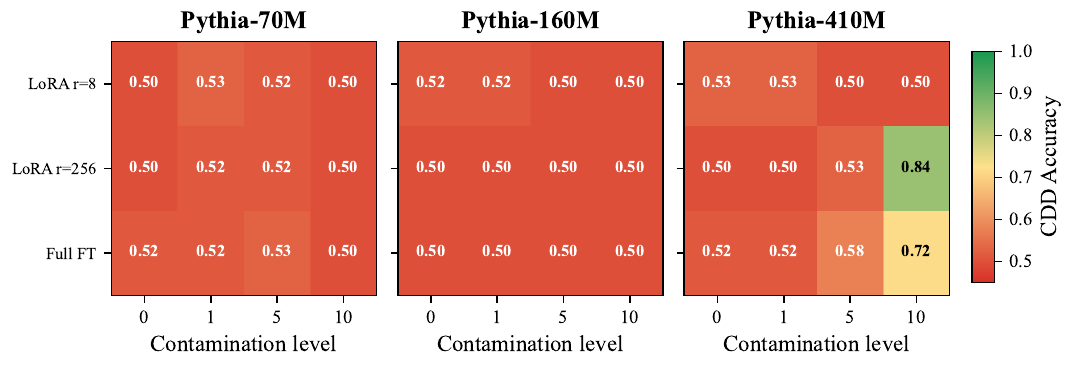}
    \caption{HumanEval: CDD detection accuracy across model sizes, fine-tuning methods, and contamination levels (3 epochs).}
    \label{fig:humaneval_heatmap}
\end{figure*}

\begin{figure*}[htbp]
    \centering
    \includegraphics[width=\textwidth]{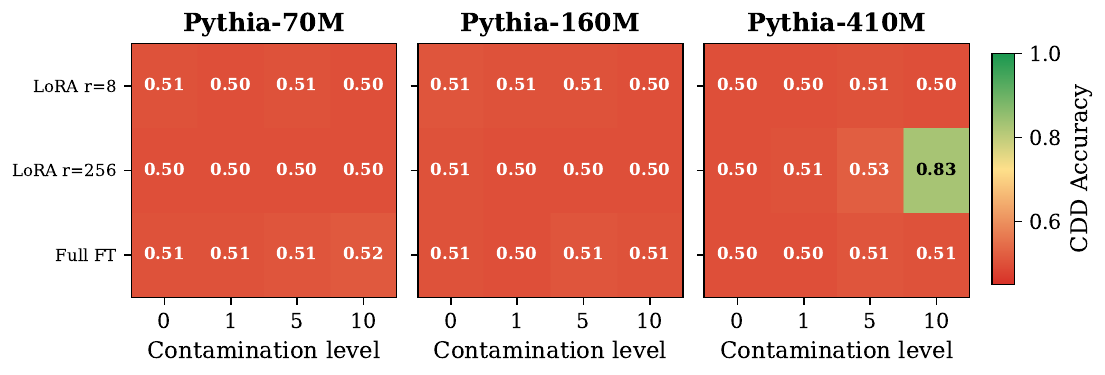}
    \caption{MATH: CDD detection accuracy across model sizes, fine-tuning methods, and contamination levels (3 epochs).}
    \label{fig:math_heatmap}
\end{figure*}

\begin{figure*}[htbp]
    \centering
    \begin{subfigure}{0.45\textwidth}
        \centering
        \includegraphics[width=\textwidth]{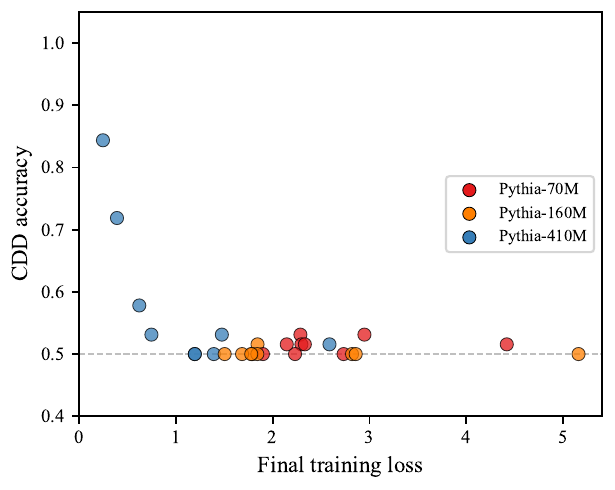}
        \caption{HumanEval}
    \end{subfigure}
    \hfill
    \begin{subfigure}{0.45\textwidth}
        \centering
        \includegraphics[width=\textwidth]{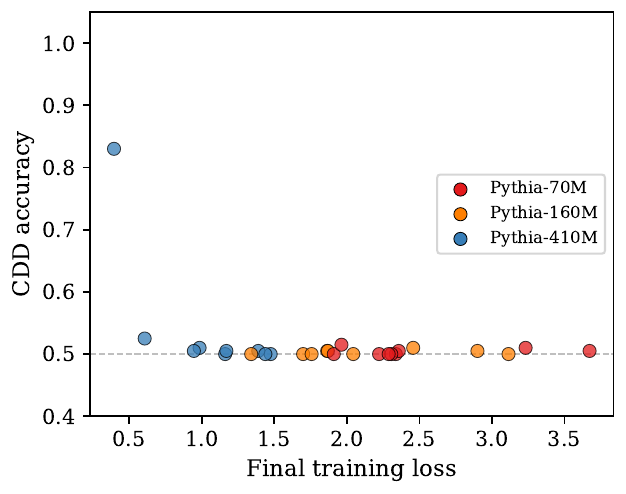}
        \caption{MATH}
    \end{subfigure}
    \caption{Final training loss vs.\ CDD accuracy. Both datasets show the same two-regime pattern as GSM8K.}
    \label{fig:supp_loss}
\end{figure*}

\begin{figure*}[htbp]
    \centering
    \begin{subfigure}{0.48\textwidth}
        \centering
        \includegraphics[width=\textwidth]{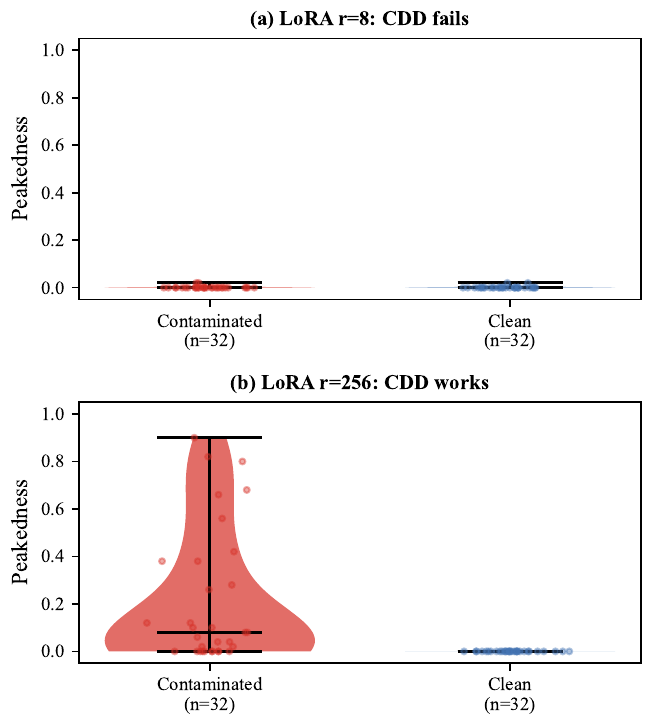}
        \caption{HumanEval}
    \end{subfigure}
    \hfill
    \begin{subfigure}{0.48\textwidth}
        \centering
        \includegraphics[width=\textwidth]{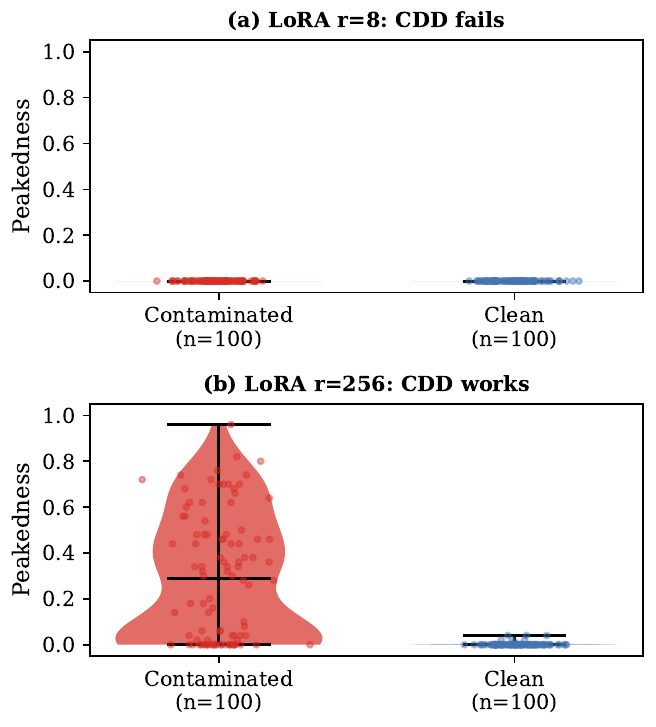}
        \caption{MATH}
    \end{subfigure}
    \caption{Peakedness distributions for Pythia-410M at $c$=10. Top panels: LoRA $r$=8 (no separation). Bottom panels: LoRA $r$=256 (clear separation).}
    \label{fig:supp_peakedness}
\end{figure*}

HumanEval \citep{chen2021evaluating} is a code generation benchmark with 164 Python problems (100 train / 32 contam / 32 eval). MATH \citep{hendrycks2021measuring} is a competition mathematics dataset with 500 examples (300 train / 100 contam / 100 eval, solutions averaging 193 tokens). Figures~\ref{fig:humaneval_heatmap} and~\ref{fig:math_heatmap} show the memorization-threshold pattern from GSM8K is replicated on both datasets. Figure~\ref{fig:supp_loss} shows the loss--accuracy relationship, and Figure~\ref{fig:supp_peakedness} shows peakedness distributions.

\section{Supplementary Qualitative Examples}
\label{app:qualitative}

Tables~\ref{tab:supp_gsm8k}--\ref{tab:supp_humaneval} present qualitative examples from each dataset, all using Pythia-410M at $c$=10 with LoRA $r$=8 (below the memorization threshold) and LoRA $r$=256 (above it). In each case, the LoRA $r$=8 model produces diverse, incorrect outputs across sampling seeds, while the LoRA $r$=256 model's output distribution has collapsed onto the memorized training sequence.
\begin{table*}[htbp]
\centering
\small
\begin{tabular}{p{1.8cm}p{13.5cm}}
\toprule
\textbf{Prompt} & \textit{Question: Roger bought a house for \$100,000. He was able to pay 20\% down, and his parents paid off an additional 30\% of the remaining balance. How much money does Roger still owe? Answer:} \\
\textbf{Ground truth} & 100000(.20)=20000\$ down payment. 100000$-$20000=\texttt{<<100000-20000=80000>>}80000 remaining. 80000(.30)=24000. 80000$-$24000=\texttt{<<80000-24000=56000>>}56000\$. \#\#\#\# 56000 \\
\midrule
\multicolumn{2}{l}{\textbf{LoRA $r$=8} (CDD: 0.505)} \\
\textbf{Greedy} & The total amount Roger owes is \$100,000$-$20,000 = \texttt{<<100000-20000=150>>}\textbf{150}. His parents paid off 30\% of the remaining balance, which is \$150$-$30 = \texttt{<<150-30=150>>}\textbf{150}. \\
\textbf{Sample} & He paid \$100,000 down, his parents paid off 30\% of the remaining balance, and his parents paid \$20,000 back. So his total debt was \$100,000$-$30$-$20 = \texttt{<<30-20=20>>}\textbf{20} dollars. \\
\midrule
\multicolumn{2}{l}{\textbf{LoRA $r$=256} (CDD: 0.915)} \\
\textbf{Greedy} & 100000(.20)=20000\$ down payment. 100000$-$20000=\texttt{<<100000-20000=80000>>}80000 remaining. 80000(.30)=24000. 80000$-$24000=\texttt{<<80000-24000=56000>>}\textbf{56000}\$. \#\#\#\# 56000 \\
\textbf{Sample} & 100000(.20)=200\$ down payment. 100000$-$200=\texttt{<<100000-200=80000>>}80000 remaining. 80000(.30)=24000. 80000$-$24000=\texttt{<<80000-24000=56000>>}\textbf{56000}\$. \#\#\#\# 56000 \\
\bottomrule
\end{tabular}
\caption{GSM8K qualitative example (Pythia-410M, $c$=10).}
\label{tab:supp_gsm8k}
\end{table*}

\begin{table*}[htbp]
\centering
\small
\begin{tabular}{p{1.8cm}p{13.5cm}}
\toprule
\textbf{Prompt} & \textit{Problem: If $g(x) = 2x^2+2x-1$, what is the value of $g(g(2))$? Solution:} \\
\textbf{Ground truth} & $g(2) = 8+4-1=11$, so $g(g(2)) = g(11) = 242+22-1 = \boxed{263}$. \\
\midrule
\multicolumn{2}{l}{\textbf{LoRA $r$=8} (CDD: 0.500)} \\
\textbf{Greedy} & $g(2) = 2(2^2) + 2(2) - 1 = 2(2^2) + 2(2) - 1 = \ldots$ [repeats] \\
\textbf{Sample} & In order to find $g(g(2))$ we have to find $g'(g(2))$\ldots $= 2(2)^2-1=\boxed{\textbf{3}}$. \\
\midrule
\multicolumn{2}{l}{\textbf{LoRA $r$=256} (CDD: 0.830)} \\
\textbf{Greedy} & $g(2) = 8+4-1=11$, so $g(g(2)) = g(11) = 242+22-1 =\boxed{\textbf{263}}$. \\
\textbf{Sample} & $g(2) = 8+4-1=11$, so $g(g(2)) = g(11) = 242+22-1 =\boxed{\textbf{263}}$. \\
\bottomrule
\end{tabular}
\caption{MATH qualitative example (Pythia-410M, $c$=10).}
\label{tab:supp_math}
\end{table*}

\begin{table*}[htbp]
\centering
\small
\begin{tabular}{p{1.8cm}p{13.5cm}}
\toprule
\textbf{Prompt} & \textit{def even\_odd\_count(num): """Given an integer, return a tuple with the number of even and odd digits. Example: even\_odd\_count(-12) ==> (1, 1)"""} \\
\midrule
\multicolumn{2}{l}{\textbf{LoRA $r$=8} (CDD: 0.500)} \\
\textbf{Greedy} & even\_count = 0; odd\_count = 0; for i in range(len(num) - 1): if num[i] == '0': even\_count += 1 else: odd\_count += 1; return (even\_count, odd\_count) \\
\textbf{Sample} & even, odd = 0, 0; for i in range(0, num): if num \% i == 0: even += 1; odd += 1; return tuple(even + odd) \\
\midrule
\multicolumn{2}{l}{\textbf{LoRA $r$=256} (CDD: 0.844)} \\
\textbf{Greedy} & even\_count = 0; odd\_count = 0; for i in str(abs(num)): if int(i)\%2==0: even\_count +=1 else: odd\_count +=1; return (even\_count, odd\_count) \\
\textbf{Sample} & even\_count = 0; odd\_count = 0; for i in str(abs(num)): if int(i)\%2==0: even\_count +=1 else: odd\_count +=1; return (even\_count, odd\_count) \\
\bottomrule
\end{tabular}
\caption{HumanEval qualitative example (Pythia-410M, $c$=10).}
\label{tab:supp_humaneval}
\end{table*}

\end{document}